\documentclass[11pt]{article}

\usepackage[preprint]{acl}

\usepackage{times}
\usepackage[table]{xcolor} 
\usepackage{booktabs}  
\usepackage{makecell}
\usepackage[table]{xcolor} 
\usepackage{booktabs}   
\usepackage{multirow}   
\usepackage{siunitx}    
\usepackage{colortbl}
\usepackage{multirow}

\usepackage{latexsym}
\definecolor{StatusGreen}{RGB}{0,115,0}
\definecolor{StatusAmber}{RGB}{220,110,0}
\definecolor{StatusRed}{RGB}{170,0,0}
\usepackage{tabularx}
\usepackage{times}
\usepackage{latexsym}
\usepackage{hyperref}
\usepackage{url}
\usepackage{multirow} 
\usepackage{xcolor}
\usepackage{graphicx}
\usepackage{multirow}
\usepackage{enumitem}
\usepackage{tabularx}
\usepackage{booktabs} 
\usepackage[most]{tcolorbox}
\usepackage{amsmath}
\usepackage{amsfonts}
\usepackage{amssymb}
\usepackage{geometry}
\usepackage{graphicx}
\usepackage{siunitx}
\usepackage{booktabs}
\usepackage{array} 
\usepackage{lscape}
\usepackage{booktabs}
\usepackage{siunitx}
\usepackage{tabularx}
\usepackage{threeparttable}
\usepackage{float}

\usepackage{pifont}

\usepackage{listings}
\usepackage{multirow} 
\usepackage[T1]{fontenc}

\usepackage[utf8]{inputenc}

\usepackage{microtype}

\usepackage{inconsolata}

\usepackage{graphicx}

\title{Training Therapeutic Judges and Multi-Agent Systems for Human-Aligned Mental Health Support}

\author{
\textbf{Mizanur Rahman}\textsuperscript{1,*},
\textbf{Abeer Badawi}\textsuperscript{1,2,3,*},
\textbf{Elahe Rahimi}\textsuperscript{2,4},
\textbf{Laleh Seyyed-Kalantari}\textsuperscript{1,2,3},\\
\textbf{Frank Rudzicz}\textsuperscript{2,4},
\textbf{Enamul Hoque}\textsuperscript{1},
\textbf{Elham Dolatabadi}\textsuperscript{1,2,3} \\
\\
\textsuperscript{1}York University, Ontario, Canada \\
\textsuperscript{2}Vector Institute, Ontario, Canada \\
\textsuperscript{3}Connected Minds, Ontario, Canada \\
\textsuperscript{4}Dalhousie University, Nova Scotia, Canada \\
\textsuperscript{*}Equal contribution
}

\begin{document}
\maketitle
\begin{abstract}
 Large language models show promise for mental health support, yet therapeutic quality improves only when evaluation functions as an actionable control signal rather than a passive metric. We introduce a framework that formulates therapeutic response generation as a decision-refinement problem driven by multi-dimensional, human-aligned evaluation. In Stage I, we introduce TheraJudge, an open-source therapeutic evaluator trained via preference-based optimization on human-annotated data to produce reliable judgments across 7 psychological dimensions. In Stage II, we introduce TheraAgent, which operationalizes TheraJudge’s evaluations through a coordinated refinement process with specialized Critic, Coach, and Therapist roles that translate evaluative signals into targeted response revisions. Empirically, TheraJudge achieves strong agreement with clinician ratings, with intraclass correlation coefficients (ICC = 0.87-0.95), surpassing supervised baselines and strong closed-source judges, particularly on critical dimensions such as Safety, Relevance, and Empathy. Acting on these evaluations, TheraAgent yields a +0.43 improvement in human-rated therapeutic quality (on a 5-point scale) under blind evaluation, with 96\% clinician inter-rater reliability. Low-quality responses ($\leq 3$) improve by +2.45 points with a 94\% recovery rate, demonstrating targeted correction of unsafe outputs. Overall, our results indicate that effective alignment of mental-health LLMs stems from acting on human-aligned evaluation, rather than relying solely on stronger generation. We release code at https://github.com/vis-nlp/TheraAlign.

\end{abstract}

\section{Introduction}

\begin{figure}[!t]
    \centering
    \includegraphics[width=1\columnwidth, height = 4.3 cm]{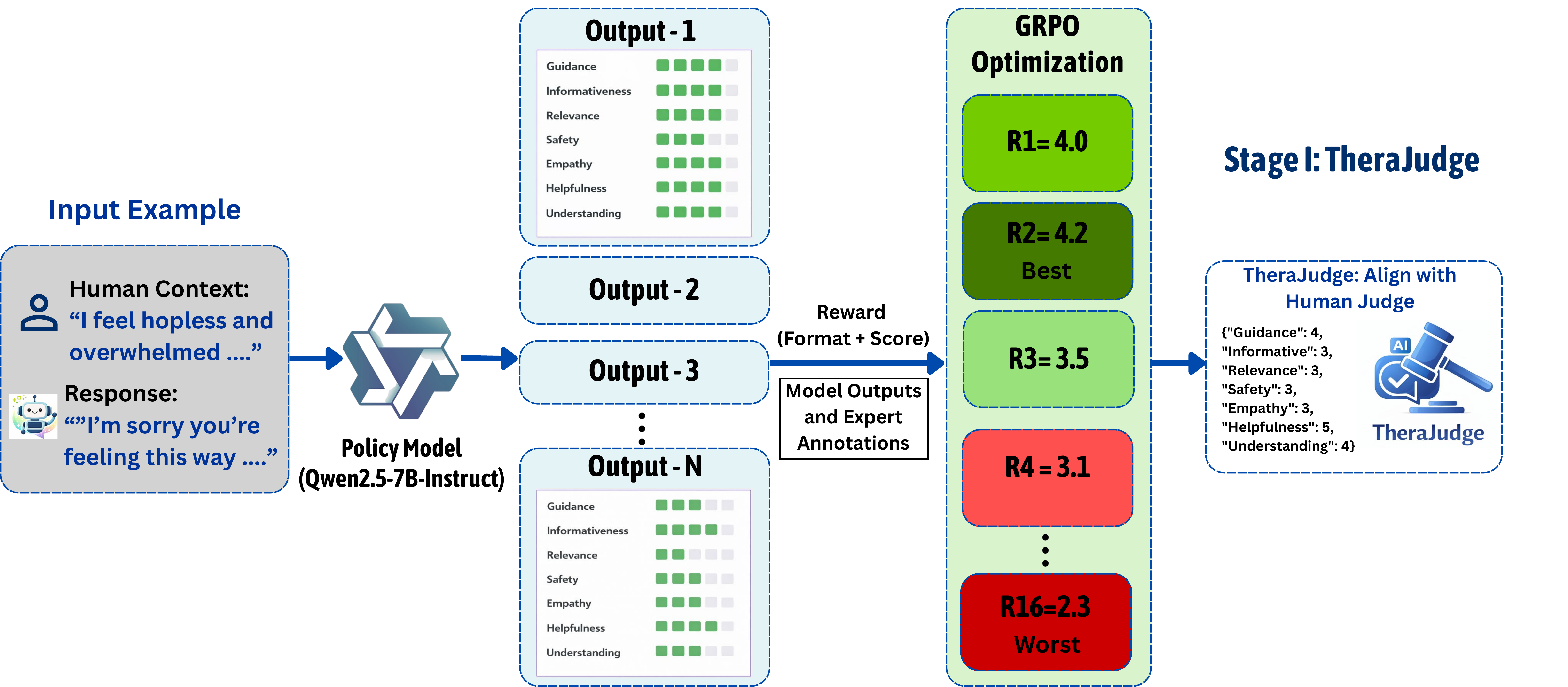}
    \caption{%
    \textbf{Stage I: TheraJudge.}
    Given a user context and response, TheraJudge applies group-wise preference optimization to produce structured multi-dimensional therapeutic ratings that guide human-aligned response refinement.
    }

    \vspace{-5mm}
\end{figure}

Recent advances in large language models (LLMs) have popularized the use of LLMs as judges for evaluating generated responses \cite{croxford2025automating, laskar2023systematic,rahman2025text2vis}. In general domains, closed-source evaluators can achieve high-quality and reliable assessments of response quality \cite{tan2024judgebench,li2025generation}. However, such evaluators are often not validated against therapeutic rubrics and safety-critical preferences. 

Mental health support requires multi-dimensional therapeutic judgment, including appropriate empathy, actionable guidance, and safe escalation. It also prioritizes open-source deployment for data governance and privacy, whereas small open-source models typically underperform in these areas, creating a critical capability gap \cite{badawi2025can}.  More importantly, evaluation by itself is not enough. Scoring, ranking, or filtering responses does not guarantee improvement unless evaluative signals are explicitly used to guide revision \cite{oh2024generative}. We refer to this limitation as the \textbf{evaluation–action} gap \cite{fernandes2023bridging}. Closing this gap is particularly important in high-stakes domains, where response quality depends on satisfying multiple, often competing objectives \cite{manna2024reconciling}.



In mental health applications, this challenge is especially acute, as models must not only remain safe and coherent but also deliver contextually appropriate, empathetic, and actionable support \cite{xu2024mental, bedi2025testing}. Therapeutic improvement therefore requires both a human-aligned evaluator that can reliably assess response quality across multiple dimensions and a mechanism that can act on those evaluations to improve responses. Although recent work has explored LLM-as-a-Judge frameworks in clinical summarization and diagnostic settings \cite{croxford2025automating, li2025generation}, a strong, open-source evaluator aligned with mental health standards remains lacking. At the same time, much of the existing work relies on proprietary, closed-source models that are neither well-aligned with therapeutic response generation nor suitable for high-stakes domains where data sensitivity and privacy constraints are critical. Moreover, such judges fail to capture human therapeutic preferences \cite{gabriel2024can, guo2024large}. Without a human-aligned and openly accessible evaluator, it is not possible to systematically improve or align models toward clinically appropriate therapeutic behavior \cite{badawi2026secureai4h}. The core challenge is therefore not only one of generation capacity, but of \textit{how evaluation is used}.

In this work, we argue that effective response improvement requires transforming evaluation into structured, explainable, and actionable \textit{signals}. Our approach differs from simple self-refinement or generic multi-agent frameworks by using clinician-aligned, multi-dimensional therapeutic evaluation as an explicit inference-time control signal for targeted response refinement. Rather than viewing response generation as a single-shot process, we frame it as decision-making under uncertainty, where evaluative signals serve as control variables. To operationalize this idea, we introduce a structured response-improvement framework that embeds a human-aligned evaluator within a refinement process explicitly designed to act on its judgments. First, we train a human-aligned therapeutic evaluator, \textit{TheraJudge}, using preference-based optimization on 10,000 conversations from the Mental-Align-100K dataset \cite{badawi2025can}. \textit{TheraJudge} produces structured ratings across multiple therapeutic dimensions, achieving high reliability and outperforming supervised baselines and closed-source evaluators. Second, we propose \textit{TheraAgent}, which utilizes these ratings to guide the refinement of targeted responses through a process of evaluation, critique, and revision. This design enables interpretable improvement when responses fall short, without relying on monolithic regeneration. We evaluate not only mean score changes, but also effect sizes, recovery rates for initially low-quality responses, and inter-rater reliability (IRR) among licensed clinicians, enabling us to assess whether refinement corrects failure modes while preserving high-quality behavior.

Across extensive experiments, we demonstrate that acting on structured evaluative feedback yields consistent improvements over judge-only and generation-only baselines. Using blind human evaluation on challenging conversations, each annotated across seven therapeutic dimensions, the overall mean score increases from 4.26 to 4.69 (+0.43). Importantly, these gains are highly selective rather than uniform. Responses that are initially low-quality or unsafe exhibit large improvements, while responses that are already acceptable remain stable and do not regress. Safety improvements, in particular, follow a thresholded pattern: although mean safety scores change little due to saturation among already-safe responses, unsafe cases are consistently corrected, achieving near-universal recovery to acceptable safety levels. These results demonstrate meaningful improvements in response quality, safety, and human alignment, underscoring the importance of closing the loop between evaluation and action. This before--and--after comparison provides an external assessment independent of the learned evaluator.

In short, our contributions are:
\textbf{(i)} We formulate therapeutic response generation as a decision-refinement problem, where multi-dimensional, human-aligned judgments serve as explicit control signals that turn assessment into actionable revision, directly addressing the evaluation–action gap; 
\textbf{(ii)} We instantiate this formulation by training an open-source therapeutic evaluator \textbf{TheraJudge} via preference-based optimization, producing interpretable, clinically meaningful ratings that enable targeted refinement beyond scalar rewards; and 
 \textbf{(iii)} We introduce \textbf{TheraAgent}, which acts on these ratings to selectively repair low-quality responses, improving the overall mean human score from 4.26 to 4.69 (+0.43) under blind evaluation and achieving high recovery rates with strong clinician inter-rater reliability (IRR).
\vspace{-2mm}

\begin{figure*}[!t]
    \centering
    \includegraphics[width=1.03\textwidth, height= 5cm]{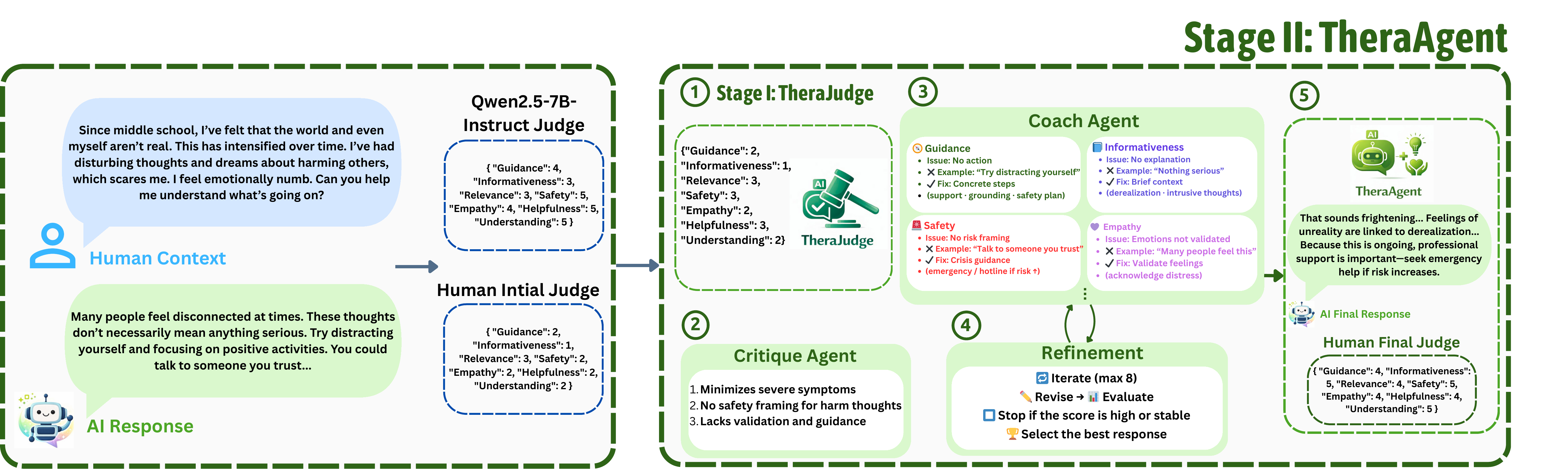}
    \caption{%
    \textbf{Stage II: TheraAgent.} Using the human-aligned scores from \textit{TheraJudge} (Stage I), TheraAgent performs critique, coaching, and refinement to transform an initial response into a final response better aligned with clinician judgment.
    }
    \vspace{-6mm}
    \label{fig:therapeutic_pipeline}
\end{figure*}

\section{Related Work}
\vspace{-10mm}
\paragraph\noindent\paragraph{\textbf{LLMs for Mental-Health Support}}  LLMs have increasingly been used in mental health for supportive communication, counseling, and empathetic dialogue generation \cite{gabriel2024can, ovsyannikova2025third, xu2025mentalchat16k,rahman2025llm}. These systems can produce coherent, emotionally attuned responses and assist with tasks such as answering queries, providing stress-reducing guidance, or triaging users for further care \cite{lai2023supporting,2026kgr, faiirv1-2024}. However, empirical evaluations reveal substantial risks, including hallucinations, inconsistent therapeutic quality, and demographic disparities in expressed empathy \cite{gabriel2024can, guo2024large}. Recent studies emphasize that, despite their potential, mental-health LLMs require structured evaluation, ethical safeguards, transparent oversight, and domain-specific fine-tuning before they can be deployed \cite{badawi2025position, ji2023rethinking, stade2024large, lawrence2024opportunities}. MentaLLaMA \cite{yang2024mentallama} attempted to match closed-source performance without expert-curated datasets and rigorous alignment. In summary, while LLMs show clear promise for mental-health support, the field lacks reliable evaluation and alignment methods needed to ensure safety and therapeutic consistency.  
\vspace{-10mm}
\paragraph\noindent\paragraph{\textbf{LLMs as Mental-Health Judges}} 
LLM-as-a-Judge has become a scalable alternative to human evaluation in open-ended language tasks. Strong models can approximate human preferences with high agreement \cite{zheng2023judging}. However, judge reliability varies considerably, and even frontier models struggle on challenging reasoning or correctness-based comparisons, motivating more rigorous evaluation frameworks \cite{tan2024judgebench} and highlighting the need for transparent, domain-specific judgment systems \cite{li2025generation}.
In healthcare, LLM-based judges have been used for clinical summarization and documentation, where automated evaluators can approximate expert ratings \cite{croxford2025automating}. However, studies reveal inconsistent evaluation practices and point to the heavy reliance on closed-source judges, which restricts transparency and auditability \cite{bedi2025testing}. Despite growing interest in LLM-as-a-Judge, no prior work provides a human-aligned open-source evaluator for therapeutic dialogue, and existing judges do not reliably capture human therapeutic preferences \cite{gabriel2024can, guo2024large}.  To address this gap, we introduce a human-aligned, open-source therapeutic evaluator trained via preference-based reinforcement learning, which provides the foundation for our multi-agent framework \textit{TheraAgent}.
\vspace{-8mm}
\paragraph\noindent\paragraph{\textbf{RL for Alignment in Mental Health}} Reinforcement learning (RL) has increasingly been used to align language models with human expectations \cite{wang2024comprehensive}. Early alignment methods such as RLHF \cite{ouyang2022training} combine human demonstrations with preference rankings and typically optimize these signals using Proximal Policy Optimization (PPO) \cite{schulman2017proximal}, which stabilizes policy updates through clipped objectives. Later methods, such as Direct Preference Optimization (DPO), simplify this pipeline by removing the need to train a separate reward model and instead optimizing the policy to match human preference ratios \cite{rafailov2023direct}. More recently, GRPO extends preference-based training by comparing groups of responses rather than pairs, yielding more stable gradients and improved sample efficiency while still avoiding the complexities of reward modeling \cite{shao2024deepseekmath, rahman2026aligning}. RL methods have also been widely explored in healthcare for treatment planning, diagnostics, and clinical decision support, illustrating their potential in high-stakes environments \cite{yu2021reinforcement, aliyu2024optimization}. Recent supportive-dialogue frameworks such as TherapyGym \cite{huangtherapygym} and Kardia-R1 \cite{yuan2025kardia} mainly use evaluation signals in RL or rubric-guided training-time alignment to directly improve generation. For human-aligned response generation, however, it is also crucial to learn evaluation signals that can be operationalized at inference time, rather than only through training-time policy updates. Table~\ref{tab:positioning_strict_binary} summarizes these distinctions between prior work and the current approach.

\vspace{-2mm}
\section{Methodology}
\vspace{-2mm}
\subsection{Problem Formulation}
We study the problem of therapeutic response improvement under multi-dimensional human preference constraints. Given a user context 
$c \in \mathcal{C}$ (a user message describing emotional or psychological concerns) and an initial model response $r_0 \in \mathcal{R}$ (produced by an LLM), the objective is to produce a refined response $r^*$ that better aligns with human therapeutic standards across multiple, potentially competing dimensions such as safety, empathy, relevance, and guidance. We formalize this task as a response refinement problem, in which improvement is driven explicitly by evaluative judgments. Central to this formulation is a human-aligned evaluator $E_{\theta}: \mathcal{C} \times \mathcal{R} \rightarrow \mathbb{R}^K$ which maps a context–response pair $(c,r)$ to a structured vector of quality signals:
\begin{equation}
E_{\theta}(c, r) = (y_1, y_2, \ldots, y_7),
\label{eq:evaluator}
\end{equation}
where each $y_i \in \{1,2,3,4,5\}$ corresponds to a rating for a therapeutic dimension (Guidance, Informativeness, Relevance, Safety, Empathy, Helpfulness, Understanding). Response improvement is governed by a refinement operator (policy)
\begin{equation}
G_{\phi}: (c, r, E_\theta(c,r)) \mapsto r^{'}
\end{equation}

which updates a candidate response by acting on its evaluative profile, targeting specific therapeutic dimensions identified by the evaluator. The refined response $r^{\ast}$ is obtained by applying the refinement operator to an initial response $r_0$:

\begin{equation}
r^{\ast} = G_{\phi}(c, r_0, E_{\theta}(c, r_0)),
\label{eq:refinement}
\end{equation}


with the objective that
\begin{equation}
E_{\theta}(c, r^{\ast}) \succ E_{\theta}(c, r_0),
\label{eq:improvement}
\end{equation}

where $\succ$ denotes improvement across at least one therapeutic dimension.

Crucially, in this formulation, evaluation provides explicit, interpretable control signals, while refinement selectively targets deficient dimensions without compromising already satisfactory dimensions. 
This enables optimization over a multi-objective therapeutic space, rather than collapsing quality into a single scalar reward.

\subsection{Base Models and Role Decoupling}

We use Qwen2.5-7B \cite{yang2025qwen2} as the base model to train our therapeutic evaluator, which we refer to as TheraJudge (see App. \ref{app:model_selection}). We select Qwen for its strong open-source performance, compatibility with preference-based alignment, and the transparency required for reproducible research in sensitive mental health settings \cite{baidal2025guardians}. In the refinement stage, we explicitly decouple evaluation from generation to prevent evaluator bias and to better reflect realistic deployment scenarios in which evaluation and generation are handled by independent components. Specifically, we use LLaMA~3.1–8B  as the therapeutic \textit{response generator}, while Qwen2.5-7B serves as the \textit{Critic} and \textit{Coach}, and \textit{TheraJudge} serves as the Evaluator, providing structured multi-dimensional feedback that guides the refinement process. This role separation ensures that improvements arise from coordinated multi-agent interaction, rather than from the internal preferences of a single model, thereby increasing both the robustness and interpretability of the overall system.


\subsection{Learning a Human-Aligned Evaluator} 

In our analysis, the evaluator $E_\theta$ should reliably approximate human therapeutic judgment across multiple dimensions. We train $E_\theta$ using preference-based optimization, leveraging grouped response comparisons to learn fine-grained distinctions in therapeutic quality. To ensure both validity and high-quality alignment with human annotations, we employ a two-component reward structure.

\paragraph{\textbf{Format reward:}} Penalizes missing keys, duplicate keys, or invalid JSON outputs. This ensures that the evaluator consistently produces well-formed seven-dimensional ratings, which is critical for training stability and for downstream use in the refinement system.

\vspace{-2mm}
\paragraph{Score-based reward:}
The score-based reward is designed to reflect the core principles of therapeutic communication by integrating the evaluator’s predictions across multiple dimensions central to mental health practice. 
Rather than optimizing a single notion of quality, the objective captures the multifaceted nature of effective therapeutic responses, jointly accounting for safety, emotional attunement, relevance to the user’s concerns, clarity of guidance, and overall helpfulness.
Formally, for a user context $c$ and a candidate response $r$, we define the therapeutic reward as

\begin{equation}
R(c, r) \;=\; \sum_{k=1}^{7} w_k \cdot y_k(c, r),
\end{equation}

where $y_k(c, r) \in \{1,2,3,4,5\}$ denotes the predicted score for the $k$-th therapeutic dimension and $w_k$ is the weight.
The weight vector is defined as
\begin{equation}
\mathbf{w} = [0.10, 0.10, 0.10, 0.20, 0.25, 0.15, 0.10]
\end{equation}
corresponding to the ordered dimensions
\textit{Guidance}, \textit{Informativeness}, \textit{Relevance}, \textit{Safety}, \textit{Empathy}, \textit{Helpfulness}, and \textit{Understanding}.
Higher weights are assigned to clinically critical dimensions, particularly \textit{Safety} and \textit{Empathy}, ensuring that the process prioritizes therapeutic appropriateness and emotional well-being, as recommended by professional therapists. We also evaluate an ablated variant with uniform weights $w_k = \frac{1}{7}$ to assess the impact of dimension weighting on alignment quality.

\paragraph{Training procedure and setup:} The evaluator is trained using group-based relative policy optimization; full training procedure and implementation details are provided in Appendix~\ref{app:evaluator_training}.



\subsection{Structured Therapeutic Response Refinement}
$G_{\phi}$ improves responses by selectively modifying content associated with low-scoring dimensions while preserving strengths elsewhere. We instantiate $G_{\phi}$ using a structured decomposition of refinement into complementary functional roles:


\noindent
\textbf{(i) TheraJudge (Evaluator).}  At the core of this design is a strong evaluator that first assesses the therapeutic quality of a response, enabling the system to identify where and how the response must improve. 
Based on this assessment, the framework coordinates the roles, which together transform evaluation into targeted, actionable revision. 
\textbf{(ii) Critic.} Identifies specific issues in the response across dimensions such as clarity, tone, personalization, and conciseness. 
\textbf{(iii) Coach.} Provides concise, actionable suggestions for dimensions rated below five, including explicit guidance on which parts of the response should be revised (e.g., when the safety score is low). 
\textbf{(iv) Therapist.} Rewrites the response by integrating the critic's identified issues and the coach's improvement guidance while maintaining overall clinical appropriateness.

Starting from an initial response, the evaluator first assesses response quality, the critic highlights concrete deficiencies, the coach offers targeted improvement directions, and the therapist produces a refined response. 
This cycle repeats for up to eight iterations, with early stopping triggered when scores stabilize or exceed a predefined threshold. 
The final output is selected as the highest-quality response obtained during the refinement process.
\vspace{-2mm}
\section{Experiments}
\subsection{Dataset}

We utilize the Mental-Align-100K dataset \cite{badawi2025can}, a large collection of mental health conversation contexts paired with model-generated therapeutic responses. For evaluator training, we focus on a subset containing 10{,}000 mental health conversations, where each instance consists of a user context and a response. To avoid information leakage, we construct the train-test split at the context level, using L2-normalized MiniLM-L6-v2 embeddings and greedy farthest-point sampling to reduce semantic overlap across partitions. Specifically, we assign 80\% of the contexts to training and reserve the remaining 20\% for testing, ensuring that no context or closely related semantic variant appears in both sets. This yields 8{,}000 annotated context-response groups for training and 2{,}000 for testing. As a result, the evaluator must generalize to unseen user contexts rather than memorizing context-specific patterns from the training data.

For therapeutic response generation, we further select 1{,}000 contexts from the Mental-Align-100K dataset. From these, we identify 267 samples covering diverse mental health issues with initially low-quality responses  for human evaluation of improvement.


\subsection{Baselines}

We compare TheraJudge against a zero-shot baseline (Qwen2.5-7B-Instruct; \cite{yang2025qwen2}), a supervised fine-tuning baseline (Qwen2.5-7B-SFT), and strong closed-source models from the previous work results, including GPT-4o \cite{achiam2023gpt}, Claude~3.7~Sonnet \cite{anthropic2024claude3}, Gemini~2.5~Flash \cite{comanici2025gemini}, and o4-mini \cite{badawi2025can}. We evaluate the quality of therapeutic responses by comparing the initial responses with the final refined responses generated by LLaMA~3.1–8B-Instruct \cite{grattafiori2024llama} using the full \textit{TheraAgent} refinement pipeline.


\subsection{Evaluation Metrics}

\paragraph{Evaluator Metrics}
Evaluator quality is assessed by measuring how closely the model’s ratings match human clinical judgments. We compute intraclass correlation coefficients (ICC) between evaluator predictions and human ratings across the seven therapeutic dimensions, providing a measure of reliability and consistency \cite{badawi2025can}. ICC captures whether the evaluator preserves the relative ordering of responses within each comparison group, which is essential for preference-based alignment and judgment fidelity. We adopt conventional ICC interpretation thresholds, where values below $0.5$ indicate poor reliability, values between $0.5$ and $0.75$ indicate moderate reliability, and values above $0.75$ indicate excellent reliability. Under this criterion, ICC directly reflects whether TheraJudge serves as a suitable proxy for human therapeutic judgment. For details see  Appendix~\ref{app:statistical_details}.

\vspace{-3mm}
\paragraph{Therapeutic Response Metrics}
Therapeutic response quality is evaluated using the same seven-dimensional rating scheme. Each generated response is scored on Guidance, Informativeness, Relevance, Safety, Empathy, Helpfulness, and Understanding using a 1--5 scale. The full therapeutic evaluation rubric and scoring anchors are provided in Table~\ref{tab:therapeutic_rubric}. We measure both per-dimension changes and an aggregated average therapeutic score. This setup enables a direct comparison of initial and refined responses, allowing us to quantify how effectively the multi-agent refinement system improves therapeutic behavior. 

\vspace{-1mm}
\subsection{Human Evaluation}

To validate our evaluation framework, we first conducted an inter-rater reliability (IRR) analysis with three independent licensed clinicians, who evaluated 267 AI-generated therapeutic responses across seven dimensions, yielding 5,607 total ratings (267 conversations $\times$ 3 evaluators $\times$ 7 dimensions ). The analysis yielded 96\% exact agreement and 100\% majority agreement on clinically meaningful quality thresholds (inadequate vs. adequate), demonstrating strong inter-rater consensus (Appendix~\ref{app:irr}). Having established the reliability of the evaluation rubric, we then conducted before–after evaluation of \textit{TheraAgent}’s refinement behavior. In this phase, one of the same clinicians blindly rated both the initial and refined responses across the seven therapeutic dimensions, together with qualitative assessments of safety, emotional appropriateness, clarity of guidance, and alignment with best therapeutic practices. This evaluation assesses whether the multi-agent refinement system produces responses that are judged by clinicians to be safer, more empathetic, and more therapeutically effective. Clinician qualifications, stage-specific roles, evaluation procedures, and statistical testing details are provided in Appendix Table~\ref{tab:clinician_details} and Appendix~\ref{app:statistical_details}.

\vspace{-2mm}
\section{Results}

\definecolor{HeaderBlue}{RGB}{0, 51, 102}      
\definecolor{GrayBenchmark}{RGB}{242, 242, 242} 
\definecolor{GoldBaseline}{RGB}{255, 252, 235}  
\definecolor{TealImproved}{RGB}{235, 252, 250}  
\definecolor{BlueHero}{RGB}{232, 242, 255}    

\begin{table*}[!tb]
\centering
\tiny
\renewcommand{\arraystretch}{1.2}
\setlength{\tabcolsep}{5pt}
\begin{tabularx}{0.8\textwidth}{l X S[table-format=1.3] l S[table-format=1.3] S[table-format=1.3]}
\toprule
\rowcolor{HeaderBlue}
\textbf{\color{white}Judge Group} & \textbf{\color{white}Dimension} & {\textbf{\color{white}ICC(C,1)}} & \textbf{\color{white}95\% CI} & {\textbf{\color{white}ICC(A,1)}} & {\textbf{\color{white}CI width}} \\
\midrule

\rowcolor{GrayBenchmark} & Guidance (o4-mini)      & 0.948 & [0.744, 0.976] & 0.786 & 0.233 \\
\rowcolor{GrayBenchmark} & Informativeness (o4-mini)& 0.918 & [0.638, 0.978] & 0.908 & 0.340 \\
\rowcolor{GrayBenchmark} & Relevance (Claude-3.7)  & 0.730 & [0.394, 0.987] & 0.743 & 0.594 \\
\rowcolor{GrayBenchmark} & Safety (Claude-3.7)     & 0.685 & [0.333, 0.961] & 0.597 & 0.628 \\
\rowcolor{GrayBenchmark} & Empathy (Claude-3.7)    & 0.906 & [0.429, 0.958] & 0.474 & 0.528 \\
\rowcolor{GrayBenchmark} & Helpfulness (Claude-3.7)& 0.900 & [0.734, 0.992] & 0.742 & 0.258 \\
\rowcolor{GrayBenchmark} \multirow{-7}{*}{\makecell{Best of Closed-Source \\ \cite{badawi2025can}}}
 & Understanding (o4-mini) & 0.871 & [0.636, 0.938] & 0.592 & 0.302 \\
\midrule

\rowcolor{GoldBaseline} & Guidance         & 0.650 & [0.450, 0.824] & 0.499 & 0.374 \\
\rowcolor{GoldBaseline} & Informativeness  & 0.802 & [0.573, 0.949] & 0.669 & 0.376 \\
\rowcolor{GoldBaseline} & Relevance        & 0.519 & [0.310, 0.602] & 0.263 & 0.292 \\
\rowcolor{GoldBaseline} & Safety           & 0.145 & [0.015, 0.441] & 0.079 & 0.427 \\
\rowcolor{GoldBaseline} & Empathy          & 0.616 & [0.164, 0.708] & 0.602 & 0.544 \\
\rowcolor{GoldBaseline} & Helpfulness      & 0.761 & [0.394, 0.887] & 0.740 & 0.493 \\
\rowcolor{GoldBaseline} \multirow{-7}{*}{Qwen-2.5-7B-ZS} & Understanding & 0.870 & [0.471, 0.924] & 0.712 & 0.453 \\
\midrule

\rowcolor{TealImproved} & Guidance         & 0.929 & [0.852, 0.966] & 0.911 & 0.114 \\
\rowcolor{TealImproved} & Informativeness  & 0.895 & [0.805, 0.983] & 0.886 & 0.178 \\
\rowcolor{TealImproved} & Relevance        & 0.760 & [0.636, 0.811] & 0.695 & 0.175 \\
\rowcolor{TealImproved} & Safety           & 0.605 & [0.429, 0.672] & 0.451 & 0.242 \\
\rowcolor{TealImproved} & Empathy          & 0.874 & [0.459, 0.978] & 0.866 & 0.518 \\
\rowcolor{TealImproved} & Helpfulness      & 0.921 & [0.820, 0.980] & 0.926 & 0.160 \\
\rowcolor{TealImproved} \multirow{-7}{*}{Qwen-2.5-7B-SFT} & Understanding & 0.803 & [0.646, 0.862] & 0.743 & 0.216 \\
\midrule

\rowcolor{BlueHero} & Guidance         & \bfseries 0.989 & [0.948, 0.998] & 0.981 & 0.050 \\
\rowcolor{BlueHero} & Informativeness  & \bfseries 0.983 & [0.918, 0.995] & 0.977 & 0.077 \\
\rowcolor{BlueHero} & Relevance        & \bfseries 0.960 & [0.900, 0.985] & 0.929 & 0.085 \\
\rowcolor{BlueHero} & Safety           & \bfseries 0.879 & [0.561, 0.958] & 0.848 & 0.901 \\
\rowcolor{BlueHero} & Empathy          & \bfseries 0.932 & [0.709, 0.979] & 0.919 & 0.270 \\
\rowcolor{BlueHero} & Helpfulness      & \bfseries 0.945 & [0.694, 0.986] & 0.934 & 0.291 \\
\rowcolor{BlueHero} \multirow{-7}{*}{\textbf{TheraJudge (Ours)}} & Understanding & \bfseries 0.960 & [0.882, 0.987] & 0.946 & 0.106 \\
\bottomrule
\end{tabularx}
\caption{ICC analysis with bootstrap CIs (self-bias removed; $N=9$ models per judge). CI width encodes precision. We report the performance of the highest-achieving closed-source models \cite{badawi2025can}, (GPT-4o, Gemini-2.5-Flash, and o4-mini) against our base model (Qwen-2.5-7B), SFT version, and the proposed \textit{TheraJudge}.}

\vspace{-7mm}
\label{tab:icc_results_corrected}
\end{table*}

\vspace{-2mm}
\subsection{Reliability of Human-Aligned Therapeutic Judgment}

We first evaluate whether TheraJudge reliably approximates human therapeutic judgment across the seven clinically grounded dimensions. Table~\ref{tab:icc_results_corrected} reports ICC comparing evaluator predictions to clinician ratings. TheraJudge achieves consistently high agreement across all dimensions, with ICC(C,1) values between 0.879 and 0.989 across dimensions and ICC(A,1) between 0.848 and 0.981, indicating excellent reliability. TheraJudge also achieves the lowest prediction error, with an average MSE of 0.67 and RMSE of 0.82, outperforming the supervised baseline (RMSE = 0.96) and the zero-shot baseline (RMSE = 1.17). The largest improvements over the supervised baseline (Qwen-2.5-7B-SFT) and the zero-shot baseline (Qwen-2.5-7B-ZS) occur in the \emph{Relevance}, \emph{Empathy}, and \emph{Understanding} dimensions, which require nuanced contextual interpretation and affective sensitivity. This pattern demonstrates that preference-based reinforcement learning substantially strengthens alignment with human therapeutic judgment, particularly along clinically salient dimensions. 
 
\begin{figure}[t]
\includegraphics[width=1\columnwidth, height = 5 cm]{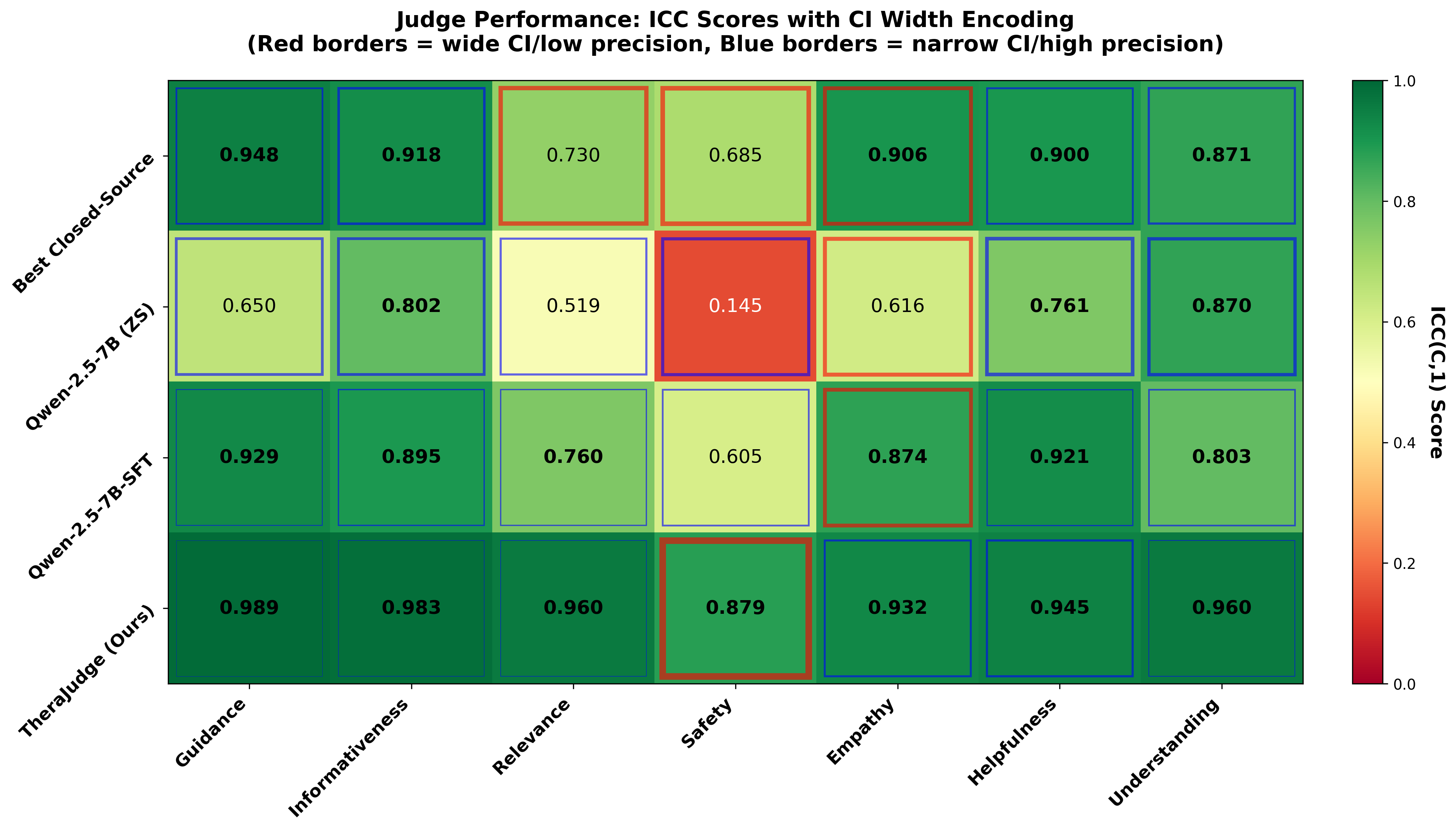}
\caption{ICC(C,1) with CI width scores across seven dimensions for four judges: Best Closed-Source, Qwen-2.5-7B ZS, Qwen-2.5-7B SFT, and TheraJudge.}
\label{fig:judge_heatmap_dual}
\vspace{-8mm}
\end{figure}

In contrast, closed-source judges—including Claude-3.7-Sonnet, GPT-4o, Gemini-2.5-Flash, and o4-mini, exhibit substantial instability on critical dimensions such as \emph{Safety} and \emph{Relevance}, despite strong performance in general-purpose language evaluation. These results highlight the limitations of non-specialized evaluators for therapeutic assessment and motivate domain-aligned training for reliable mental-health evaluation. Overall, TheraJudge establishes a new state of the art for open-source therapeutic evaluation, achieving superior reliability, stability, and balance across both cognitive and affective dimensions. As shown in Figure~\ref{fig:judge_heatmap_dual}, TheraJudge attains the highest ICC(C,1) across all seven dimensions (0.879–0.989). Additionally, we assess the impact of preference group size on evaluator alignment and train GRPO models with groups of 4, 8, and 16 responses per context, concluding that 16 provides the best results (Appendix~\ref{app:model_selection}). 


\begin{table*}[t]
\centering
\small
\renewcommand{\arraystretch}{1.15}
\setlength{\tabcolsep}{6pt}
\vspace{-6pt}
\begin{tabular}{lcccccc}
\toprule
\textbf{Dimension} &
\textbf{Initial} &
\textbf{Final} &
\textbf{$\Delta$ Improvement} &
\textbf{Cohen’s $d$} &
\textbf{Recovery Rate (\%)} &
\textbf{Low-Quality Rate (\%)} \\
\midrule
Guidance        & 3.77 & 4.17 & +0.40 & 0.36 & 76.2 & 31.5 \\
Informativeness & 3.94 & 4.57 & +0.63 & 0.58 & 95.9 & 18.4 \\
Relevance       & 4.68 & 4.96 & +0.28 & 0.62 & 100 & 15.4 \\
Safety          & 4.95 & 4.98 & +0.03 & 0.21 & 100 & 10.1 \\
Empathy         & 3.97 & 4.54 & +0.57 & 0.49 & 96.4 & 20.6 \\
Helpfulness     & 3.89 & 4.62 & +0.73 & 0.61 & 97.4 & 14.2 \\
Understanding   & 4.63 & 4.97 & +0.34 & 0.62 & 94.7 & 14.2 \\
\midrule
\textbf{Overall Mean} & \textbf{4.26} & \textbf{4.69} & \textbf{+0.43} & \textbf{—} & \textbf{94.0} & \textbf{17.9} \\
\bottomrule
\end{tabular}
\vspace{-6pt}
\caption{Human evaluation across seven therapeutic dimensions ($n=267$). Initial and Final are mean 5-point ratings before and after refinement; $\Delta$ is the mean change (Final $-$ Initial). Cohen’s $d$ denotes effect size, Low-Quality Rate the share of responses initially rated $\leq 3$, and Recovery Rate the share of those improved to $>3$. All improvements are significant ($p<.001$, paired $t$-test).}
\vspace{-7mm}

\label{tab:unified_results}
\end{table*}

\vspace{-2mm}
\subsection{Acting on Evaluation Improves Therapeutic Quality}
\label{sec:acting_on_eval}

We next examine whether structured evaluation leads to meaningful improvement when it is explicitly acted upon.   Table~\ref{tab:unified_results} shows that across all responses, acting on evaluation yields a statistically significant mean improvement of +0.43 points (4.26 $\rightarrow$ 4.69, $p<.001$, paired $t$-test).  Improvements are observed across all dimensions, with the largest gains in Helpfulness, Informativeness, and Empathy. 
Effect sizes range from small (Guidance, $d=0.36$) to medium (Relevance, $d=0.62$), indicating consistent and clinically meaningful improvements. 

Although the overall mean improvement is +0.43, this average is compressed because, even within these challenging evaluation cases, many dimension-level ratings already begin at high scores, leaving limited room for further improvement. Improvements are therefore highly selective rather than uniform. Low-quality responses (initial score $\leq 3$) improve by +2.45 points on average (110\% gain), yielding much larger gains than the overall mean. Among initially deficient cases, Safety shows the strongest transformation, improving by +3.19 points and elevating unsafe responses from 1.74/5 to 4.93/5. Correspondingly, recovery rates are near-universal: Safety and Relevance achieve 100\% recovery, while all other dimensions exceed 94\% recovery except Guidance (76.2\%), which remains the most challenging dimension in ambiguous or crisis-related contexts. Overall, 94.0\% of initially low-quality ratings are successfully elevated above the acceptable threshold. These gains are not limited to initially low-quality cases: high-scoring responses also improve despite limited headroom, suggesting that TheraAgent can refine acceptable answers as well as repair deficient ones.

To assess rating consistency, two additional independent clinical experts evaluated the same refined responses. This IRR analysis achieves 96\% exact agreement and 100\% majority agreement (Table~\ref{tab:irr_attributes}), confirming that the observed improvements are supported by blinded expert judgment and strong cross-rater consensus. All improvements are evaluated using two-tailed paired $t$-tests ($\alpha=0.007$, $p<.001$).

\vspace{-4mm}
\paragraph{Generalizability Beyond Therapeutic Dialogue.}
Although our empirical validation focuses on therapeutic mental-health dialogue, the proposed framework is not inherently specific to this domain. Its core design separates evaluation from generation by learning dimension-level quality signals from expert supervision and then using those signals to guide targeted refinement. In principle, this framework could be adapted to other domains by replacing the therapeutic rubric and supervision with domain-appropriate expert criteria. In addition, our training and evaluation data span diverse therapeutic interactions drawn from multiple sources, reducing dependence on any single dataset style. Finally, evaluation is conducted on held-out, previously unseen contexts, indicating generalization to new dialogue situations within the therapeutic domain.

\subsection{Component Contributions}

We conduct ablation studies to analyze the impact of our collaborative multi-agent design compared to simpler actor–critic style refinement. 
Removing the evaluator and relying only on the Critic and Coach reduces the overall improvement by 0.21, demonstrating the central role of explicit quality assessment in guiding effective refinement. 
Removing the Coach reduces the improvement by 0.14, and removing both the Critic and Coach simultaneously results in a larger degradation of 0.17 compared to the full system.GPT-4o is used only for controlled relative comparisons under a fixed evaluation setup, while clinician ratings validate the end-to-end therapeutic improvements. 
These results demonstrate that each agent contributes meaningfully to refinement quality and that the proposed multi-agent collaboration is essential for achieving maximal therapeutic improvement. 
We further compare a Qwen-only pipeline against a mixed pipeline with LLaMA as the generator and Qwen as the evaluator, and find that using different model families for generation and evaluation yields stronger performance than a homogeneous pipeline, suggesting that separating generation and evaluation reduces reward bias.

\vspace{-4mm}

\medskip
\vspace{-2mm}
\section{Conclusion}

In this work, we present a framework aimed at generating human-aligned therapeutic responses, and  demonstrate that achieving this goal in an agentic refinement system critically depends on the availability of a robust, human-aligned evaluator. TheraJudge provides this foundation by reliably identifying weak responses and exposing fine-grained deficiencies across clinically meaningful dimensions, enabling the Critic and Coach to produce structured feedback that directly guides improvement. This evaluation-centered design enables the multi-agent framework TheraAgent, comprising Evaluator, Critic, Coach, and Therapist, to convert judgment into targeted and interpretable refinement, resulting in consistent gains in therapeutic quality across both automatic and clinician-based evaluations. Our results demonstrate substantial improvements in Helpfulness, Informativeness, Empathy, and Guidance while maintaining high levels of Safety and Understanding, establishing a transparent and practically deployable framework for scalable human-aligned mental-health AI systems.

\paragraph{Limitations.}
This work is subject to several limitations inherent to mental-health research. Data availability in this domain is constrained by privacy, ethical, and legal considerations, which restrict both the scale and diversity of publicly accessible resources and may limit generalizability. Moreover, closed-source models often benefit from access to substantially larger and less transparent training corpora, creating an uneven comparison with open-source models; while our approach seeks to mitigate this gap through improved evaluation and alignment, it cannot fully offset differences in underlying data exposure. Finally, human evaluation in mental-health settings requires domain expertise and careful oversight, making it difficult to obtain at a large scale and limiting the breadth of expert-annotated assessments despite their importance for reliable evaluation \cite{ravenda2025are, baidal2025guardians}.

\paragraph{Ethical Considerations.}
This study received approval from the Research Ethics Board (REB). All data used is publicly available and fully anonymized, with no access to personally identifiable information. Human evaluators and automated models interacted only with de-identified text. The dataset comprises real counseling-style dialogues from clinical and online sources, supplemented by limited machine-rephrased text that does not introduce new human-authored content, as per the original dataset documentation. The evaluated systems are intended solely for research and evaluation purposes and are not designed to replace mental health professionals. We caution against deployment without human oversight. To mitigate risks related to bias or over-reliance on AI judgments, we employed a transparent evaluation pipeline, reported reliability with confidence intervals, and controlled for evaluator self-preference bias. Finally, we used
AI-based writing assistants only to improve the
presentation of the paper.

\bibliography{custom}

\appendix




\appendix

\section{Evaluator Training Procedure and Setup Details}
\label{app:evaluator_training}

During training, each context is associated with a group of 16 sampled candidate completions from the current policy, which are compared using rubric-based reward signals for relative preference optimization. The optimization procedure compares responses within each group and increases the likelihood of higher-rated responses, enabling the evaluator to learn fine-grained distinctions in therapeutic quality. We train the evaluator on two H100 GPUs using a maximum context length of 1536 tokens. Training is completed in a single epoch over the annotated dataset and converges reliably within approximately two hours, producing an evaluator that exhibits strong agreement with human therapeutic judgments.

\section{Additional Statistical Details}
\label{app:statistical_details}

This appendix provides additional details for the two statistical components used in our evaluation. First, we explain the paired \textit{t}-test used to assess whether therapeutic response quality improves after refinement. Second, for completeness, we briefly summarize the Intraclass Correlation Coefficient (ICC) analysis used to assess evaluator reliability proposed in \citep{badawi2025can}.

\subsection{Paired \textit{t}-Test for Therapeutic Improvement}
\label{app:paired_t_test}

To test whether the refinement process improves therapeutic response quality, we compare the initial response and the final refined response for the same conversation context. This yields a naturally paired design, since each context is evaluated twice under the same seven-dimensional therapeutic rubric: once before refinement and once after refinement.

Let \(x_i^{(\mathrm{init})}\) denote the score of the initial response for context \(i\), and let \(x_i^{(\mathrm{final})}\) denote the score of the refined response for the same context. For each context, we define the paired difference
\[
d_i = x_i^{(\mathrm{final})} - x_i^{(\mathrm{init})}.
\]
The null and alternative hypotheses are:
\[
H_0: \mu_d = 0
\qquad \text{vs.} \qquad
H_1: \mu_d \neq 0,
\]
where \(\mu_d\) is the population mean of the paired differences.

The paired \textit{t}-statistic is computed as
\[
t = \frac{\bar{d}}{s_d / \sqrt{n}},
\]
where \(\bar{d}\) is the sample mean of the paired differences, \(s_d\) is the sample standard deviation of those differences, and \(n\) is the number of paired observations.

In our human evaluation, \(n = 267\) conversation contexts, so the test is performed with \(df = 266\). This paired design is preferable to an independent-samples comparison because each conversation serves as its own control. As a result, the analysis controls for context-specific difficulty, emotional severity, and ambiguity, and directly tests whether refinement improves scores for the same underlying case.

We apply the paired \textit{t}-test independently to each of the seven therapeutic dimensions:
\textit{Guidance, Informativeness, Relevance, Safety, Empathy, Helpfulness,} and \textit{Understanding}. Because multiple hypothesis tests are conducted across dimensions, we use a Bonferroni-corrected significance threshold. Accordingly, dimension-level improvements are evaluated using a two-tailed threshold of \(\alpha = 0.007\). In the main paper, all reported improvements satisfy this corrected threshold.

This test answers the central question of the refinement analysis: after holding the conversation context fixed, does the final response receive a significantly higher therapeutic score than the initial response? In this sense, the paired \textit{t}-test provides direct statistical evidence that acting on evaluation leads to measurable gains in therapeutic quality.

\subsection{Intraclass Correlation Coefficient (ICC) for Realibility}
\label{app:icc_details}

ICC is used to quantify agreement between automated evaluator scores and human ratings across the seven therapeutic dimensions. It is appropriate in this setting because evaluator reliability is not only about whether two judges are positively  correlated, but also whether they preserve the same relative ordering of responses and whether they operate on compatible scoring scales.

OpenAlign adopts the ICC-based reliability analysis proposed in  \citep{badawi2025can}, where the methodological motivation, interpretive framework,  and diagnostic use of ICC were presented in detail. In OpenAlign, ICC is used as  part of the evaluator validation stage, where model-based scores are compared against human ratings to determine whether the evaluator provides sufficiently  reliable signals for refinement and downstream analysis. More specifically, ICC is used to examine agreement between automated and human evaluation at the  dimension level, allowing us to assess whether the evaluator captures human preferences in a stable and interpretable way. This is important in our setting because OpenAlign relies on evaluator feedback not only to score responses, but also to guide iterative refinement of therapeutic quality.

Following \citet{badawi2025can}, we report both consistency-based and absolute-agreement ICC results, together with bootstrap-based 95\% confidence intervals. The consistency results indicate whether the evaluator preserves the relative ordering of responses similarly to human judgment, while the absolute-agreement results reflect closer alignment in assigned score values.
\section{Base Model Selection}
\label{app:model_selection}

Because preference-based RL benefits from a capable zero-shot initializer while remaining computationally efficient, we focus on models in the 7--8B range. Under an identical training pipeline and development evaluation protocol, we compared Qwen2.5-7B, Qwen3-8B, LLaMA~3.1--8B \cite{grattafiori2024llama}, and Mistral~2.5--7B. Qwen2.5-7B achieved the strongest overall evaluator performance in our setting and also showed consistently strong behavior on mental-health-oriented conversations, so we adopt it as our default base model. LLaMA~3.1--8B performed comparably, and when we train \textit{TheraJudge} with LLaMA~3.1--8B as the policy model, we observe similar improvements, suggesting that our approach is robust across model families.


\subsection{Complete Judge Performance Results}
\label{app:complete_judge_performance}

We present a comprehensive comparison of TheraJudge against all baseline judges, including zero-shot, supervised fine-tuning (SFT), and the strongest closed-source models. Table~\ref{tab:model_summary} shows the complete performance matrix across all seven therapeutic dimensions  for seven distinct model configurations: four untrained proprietary models (Claude-3.7-Sonnet, GPT-4o, Gemini-2.5-Flash, o4-mini), and three Qwen-2.5-7B variants representing progressive training stages (Zero-shot, SFT, GRPO5/gen 16).

\subsubsection{Key Findings from Seven-Model Comparison}

\textbf{TheraJudge Dominance.} TheraJudge (gen 16) achieves the highest overall average ICC(C,1) of 0.949, substantially outperforming all baselines including the best closed-source model (Claude-3.7-Sonnet: 0.830). It wins 6 out of 7 dimension-level comparisons and is the only model to achieve excellent-level reliability (ICC > 0.75) across 6 of 7 dimensions , with only Safety rated as ``Poor'' due to wide confidence intervals despite a strong ICC value of 0.879.

\textbf{Training Impact.} Training transforms Qwen-2.5-7B from a weak baseline (Zero-shot: 0.623, ranked 6th) through supervised fine-tuning (SFT: 0.827, ranked 2nd) to state-of-the-art performance (TheraJudge gen 16: 0.949, ranked 1st). This represents a 52\% overall improvement and demonstrates that preference-based reinforcement learning substantially enhances therapeutic alignment.

\textbf{Closed-Source Limitations.} Despite their general capabilities, closed-source models exhibit critical weaknesses on therapeutic evaluation. Claude-3.7-Sonnet, the strongest closed-source judge, achieves only 0.685 on Safety and 0.730 on Relevance---both below the 0.75 excellent threshold. GPT-4o (0.480 Safety), o4-mini (0.259 Safety), and Gemini-2.5-Flash (0.306 Relevance, 0.377 Safety) perform even worse on these critical dimensions, highlighting their unsuitability for therapeutic assessment without domain-specific training.

\textbf{Safety as Critical Discriminator.} Safety evaluation proves most challenging across all models. TheraJudge (0.879) substantially outperforms all baselines, including the best closed-source model Claude (0.685, +28\% advantage) and the zero-shot baseline (0.145, +506\% improvement). This gap underscores that Safety assessment requires explicit therapeutic training and cannot be reliably performed by general-purpose models.

\begin{table}[t]
\centering
\small
\renewcommand{\arraystretch}{1.1}
\setlength{\tabcolsep}{8pt}
\begin{tabular}{@{}lcc@{}}
\toprule
\textbf{Model} & \textbf{Avg ICC} & \textbf{Rank} \\
\midrule
TheraJudge (Gen 16) & 0.949 & 1st \\
TheraJudge (Gen 4)              & 0.924 & 2nd \\
TheraJudge (Gen 8)               & 0.900 & 3rd \\
Claude-3.7          & 0.830 & 4th \\
Qwen-SFT                 & 0.827 & 5th \\
GPT-4o              & 0.739 & 6th \\
o4-mini             & 0.727 & 7th \\
Qwen-Zero-shot           & 0.623 & 8th \\
Gemini-2.5          & 0.621 & 9th \\
\bottomrule
\end{tabular}
\caption{\small{Summary of evaluated models ranked by average intraclass correlation coefficient (ICC(C,1)) with human judgments.}}
\label{tab:model_summary}
\end{table}

\subsection{Training Progression Analysis}
\label{app:training_progression}

\begin{figure*}[t]
\centering
\includegraphics[width=1\textwidth]{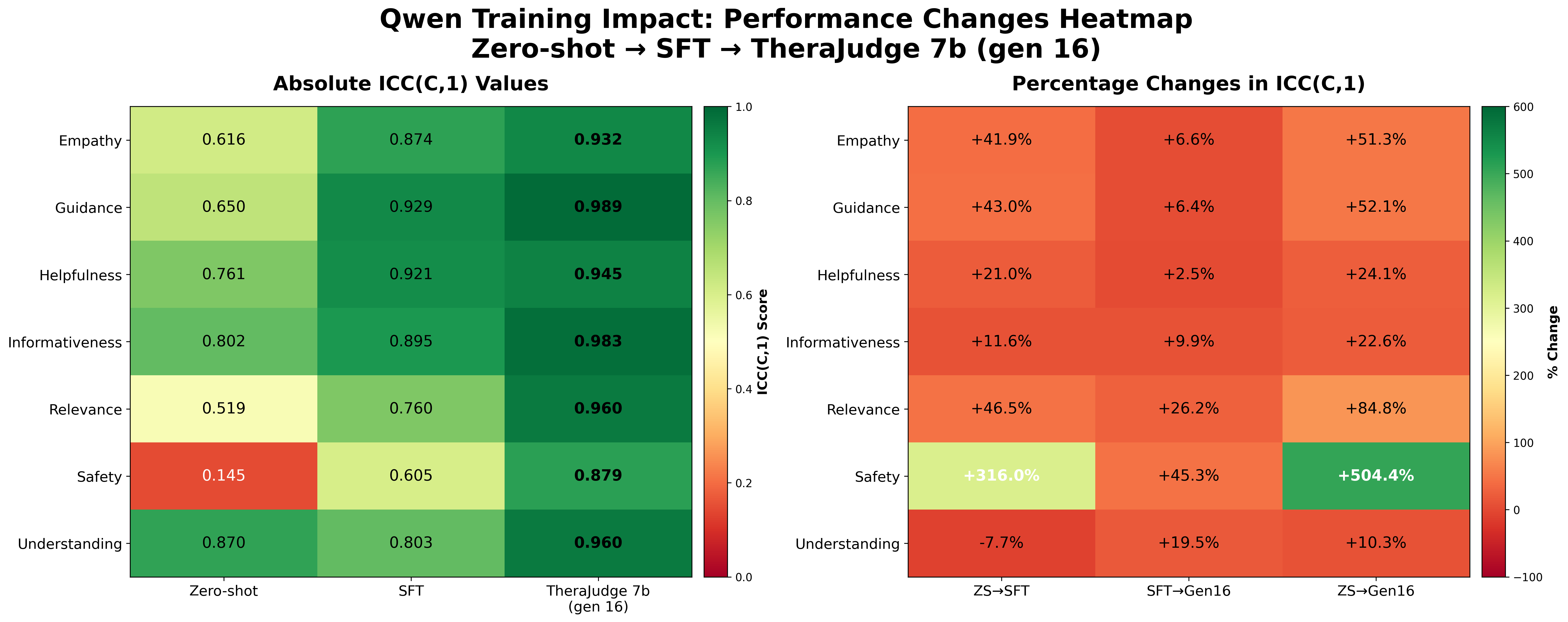}
\caption{Training impact heatmap showing absolute ICC(C,1) values (left) and percentage changes (right) across the complete training progression: Zero-shot → SFT → TheraJudge (gen 16). Safety exhibits the largest improvement (+505\%), while all dimensions  show positive gains from zero-shot to final model.}
\label{fig:training_heatmap}
\end{figure*}

We document TheraJudge's complete training trajectory from zero-shot baseline through supervised fine-tuning to Group Relative Policy Optimization (GRPO). Figure~\ref{fig:training_heatmap} visualizes the progressive improvements across all seven therapeutic dimensions.

\subsubsection{Zero-shot to SFT Progression}

Supervised fine-tuning yields substantial improvements across most therapeutic dimensions, with an average gain of +67.5\% in ICC(C,1) reliability. The largest improvements occur in Safety (+316\%), Relevance (+46.5\%), Guidance (+43\%), and Empathy (+41.9\%). However, SFT introduces a notable regression in Understanding (-7.7\%), suggesting that supervised learning alone may not capture the full complexity of human therapeutic judgment. This U-shaped pattern motivates the need for preference-based optimization.

\subsubsection{SFT to TheraJudge (gen 16) Progression}

Group Relative Policy Optimization further enhances evaluator reliability, achieving an additional +16.6\% average improvement beyond SFT. The most substantial gains occur in Understanding (+19.5\%, recovering from the SFT dip), Safety (+45.3\%, continued improvement), and Relevance (+26.2\%, substantial additional gain). Critically, GRPO eliminates the Understanding regression observed in SFT and achieves universal positive changes across all seven dimensions .

\subsubsection{Overall Training Impact}

From zero-shot baseline to final model (gen 16), TheraJudge achieves:
\begin{itemize}
\item \textbf{Safety:} 0.145 → 0.879 (+505\% improvement, largest gain)
\item \textbf{Relevance:} 0.519 → 0.960 (+85\% improvement)
\item \textbf{Guidance:} 0.650 → 0.989 (+52\% improvement)
\item \textbf{Empathy:} 0.616 → 0.932 (+51\% improvement)
\item \textbf{Informativeness:} 0.802 → 0.983 (+23\% improvement)
\item \textbf{Helpfulness:} 0.761 → 0.945 (+24\% improvement)
\item \textbf{Understanding:} 0.870 → 0.960 (+10\% improvement)
\end{itemize}

The average improvement of +107.1\% demonstrates that the complete training pipeline transforms Qwen-2.5-7B from a poor evaluator into an expert-level therapeutic judge that surpasses both supervised baselines and proprietary models. 

In addition, to assess the impact of preference-group size on evaluator alignment, we train GRPO models with groups of 4, 8, and 16 responses per context (see Table~\ref{tab:grpo_iterations_comparison}). 
Larger groups provide richer comparative signals, and we observe monotonic improvements in evaluator reliability as group size increases. 
The group-size 16 model achieves the strongest alignment with human ratings, suggesting that increased preference diversity strengthens GRPO optimization. 
These findings indicate that further scaling of preference groups is likely to yield even more accurate and robust evaluators. Key Observations: 

\begin{itemize}
\item \textbf{Monotonic improvement with group size.} Average ICC(C,1) increases consistently: gen 4 (0.924) → gen 8 (0.900, temporary dip) → gen 16 (0.949, +2.8\% from gen 4).

\item \textbf{Gen 8 represents a temporary training valley.} Performance temporarily degrades at generation 8, particularly on Safety (-6.8\%) and Empathy (-8.7\%), before recovering strongly at generation 16. This U-shaped pattern suggests that GRPO optimization may traverse local minima before converging to superior solutions.

\item \textbf{Gen 16 achieves universal excellence.} The 16-candidate configuration produces the only model with 6/7 excellent-status dimensions  and substantially tighter confidence intervals (0.254 vs 0.416 for gen 4), indicating both higher reliability and more stable performance.

\item \textbf{Confidence interval narrowing.} CI width decreases dramatically from gen 4 to gen 16 for most dimensions : Understanding: 0.570 → 0.106 (-81\%), Empathy: 0.522 → 0.270 (-48\%),  Helpfulness: 0.552 → 0.291 (-47\%), and Guidance: 0.113 → 0.050 (-56\%)

\item \textbf{Safety remains most challenging.} Despite achieving excellent ICC values across all configurations (gen 4: 0.896, gen 8: 0.835, gen 16: 0.879), Safety consistently exhibits wide confidence intervals (>0.90), reflecting inherent evaluation difficulty for this critical dimension. Gen 4 retains the best Safety score (0.896), suggesting potential trade-offs in
\end{itemize}

\section{Inter-Rater Reliability Analysis}
\label{app:irr}

To evaluate the reliability of human expert assessments of AI-generated therapeutic responses, we conducted an inter-rater reliability (IRR) analysis across three independent expert evaluators. This analysis examined consensus on a clinically meaningful binary categorization: inadequate responses (ratings 1--2) versus adequate or better responses (ratings 3--5). This threshold represents a critical distinction in mental health applications, where responses must meet minimum safety and quality standards to be suitable for deployment in real-world clinical settings.

\subsection{Methods}

\subsubsection{Sample}

The evaluation corpus consisted of 267 AI-generated therapeutic responses produced by our GRPO-optimized model. These responses represented a diverse range of mental health scenarios, including crisis support, emotional distress, relationship issues, and general well-being concerns. Each response was generated in reply to authentic user messages from mental health support contexts.

\subsubsection{Evaluators}

Three independent expert raters with clinical mental health backgrounds conducted the evaluations. All evaluators had professional experience in therapeutic communication and were trained on the evaluation framework prior to beginning the rating process. Evaluators worked independently without access to other raters' assessments to ensure unbiased judgments. Each response was evaluated across seven therapeutic dimensions.

\subsubsection{Rating Scale and Categorization}

Evaluators rated each dimension on a 5-point Likert scale (1 = very poor to 5 = excellent). For the IRR analysis, ratings were dichotomized into two clinically meaningful categories:

\begin{itemize}
    \item \textbf{Inadequate}: Ratings of 1 or 2, indicating responses that fail to meet minimum quality standards
    \item \textbf{Adequate or Better}: Ratings of 3, 4, or 5, indicating responses that meet or exceed acceptable quality thresholds
\end{itemize}

This binary categorization reflects the most critical clinical decision: whether a response is safe and appropriate for use in mental health support contexts.

\subsubsection{Statistical Measures}

Two primary metrics were used to assess inter-rater reliability:

\begin{itemize}
    \item \textbf{Exact Agreement}: The percentage of samples where all three raters assigned the same category (both inadequate or both adequate). This represents the strongest form of consensus.
    \item \textbf{Majority Agreement}: The percentage of samples where at least two of the three raters agreed on the category. This indicates reliable consensus even when one rater diverges.
\end{itemize}

\subsection{Results}

\subsubsection{Overall Inter-Rater Reliability}

The analysis revealed strong inter-rater consensus across all therapeutic dimensions. Aggregating across all seven dimensions, the three evaluators achieved the reliability metrics shown in Table~\ref{tab:irr_overall}.

\begin{table}[h]
\centering
\caption{Overall inter-rater reliability metrics across all therapeutic dimensions ($N = 267$)}
\label{tab:irr_overall}
\begin{tabular}{lc}
\toprule
\textbf{Reliability Metric} & \textbf{Value} \\
\midrule
Mean Exact Agreement (All 3 Raters) & 96.0\% \\
Majority Agreement ($\geq$2 Raters) & 100\% \\
\bottomrule
\end{tabular}
\end{table}

The 96\% exact agreement indicates that in the vast majority of cases, all three evaluators independently arrived at the same classification of response quality. The 100\% majority agreement demonstrates that even in the 4\% of cases where complete consensus was not achieved, at least two evaluators consistently agreed on the appropriate category, indicating robust reliability in quality threshold determinations.

\subsubsection{Dimension-Level Reliability}

Table~\ref{tab:irr_attributes} presents the exact agreement rates for each therapeutic dimension. Agreement was consistently high across all dimensions, with Safety demonstrating the highest consensus (98.84\%) and Guidance showing the lowest, though still strong, agreement (91.57\%).

\begin{table}[h]
\centering
\caption{Exact agreement percentages by therapeutic dimension ($N = 267$)}
\label{tab:irr_attributes}
\begin{tabular}{lc}
\toprule
\textbf{Dimension} & \textbf{Exact Agreement (\%)} \\
\midrule
Guidance & 91.57 \\
Informativeness & 96.14 \\
Relevance & 96.53 \\
Safety & 98.84 \\
Empathy & 95.37 \\
Helpfulness & 96.53 \\
Understanding & 97.27 \\
\midrule
\textbf{Mean} & \textbf{96.03} \\
\bottomrule
\end{tabular}
\end{table}

\begin{figure}[h]
\centering
\includegraphics[width=1\columnwidth]{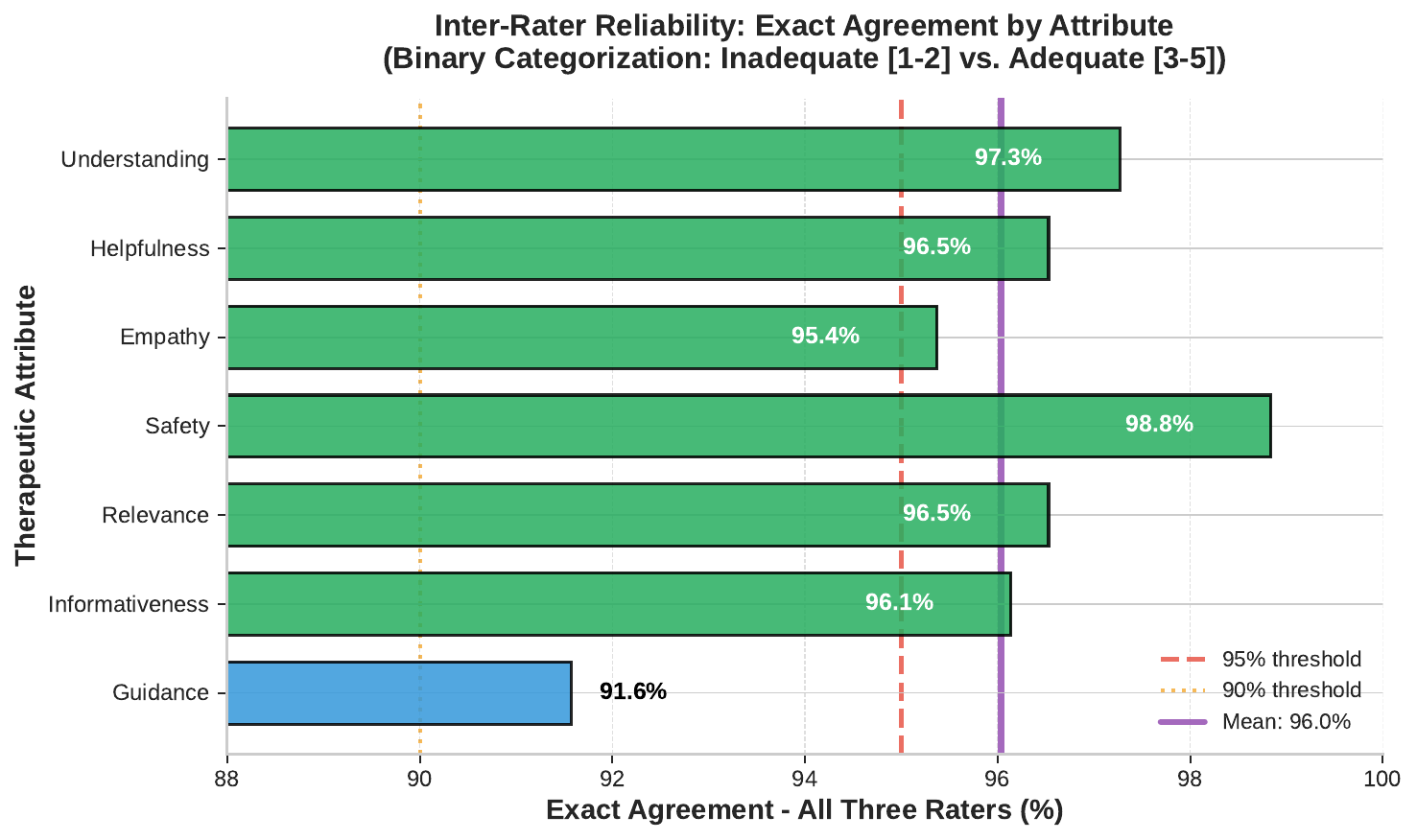}
\caption{Inter-rater exact agreement by therapeutic dimension. Bars represent the percentage of samples where all three raters agreed on the binary categorization (inadequate vs.\ adequate). Dashed lines indicate 95\% and 90\% agreement. The purple vertical line shows the mean exact agreement across all dimensions (96.0\%).}
\label{fig:irr_agreement}
\end{figure}

Figure~\ref{fig:irr_agreement} provides a visual representation of the exact agreement rates across dimensions. The consistently high agreement levels, with all dimensions exceeding 90\%, demonstrate reliable inter-rater consensus on quality categorizations across the full spectrum of therapeutic dimensions.

\subsection{Interpretation}

The 96\% mean exact agreement demonstrates strong inter-rater consensus among the three independent expert evaluators on the clinically critical distinction between inadequate and adequate therapeutic responses. This high level of agreement indicates several important findings:

\begin{enumerate}
    \item The evaluators demonstrated consistent application of quality standards across the full corpus of AI-generated responses. The ability to reliably distinguish between responses that meet or fail to meet minimum thresholds suggests that the evaluation framework captures meaningful and observable quality differences.
    
    \item The 100\% majority agreement across all 267 samples validates the robustness of quality assessments. While a complete three-way consensus was achieved in 96\% of cases, the remaining 4\% of samples showed consistent agreement between at least two evaluators. This pattern suggests that disagreements occurred primarily on genuine boundary cases rather than systematic differences in rating standards.
    
    \item The particularly high agreement on Safety (98.84\%) is especially noteworthy for clinical applications. This near-perfect consensus on identifying safe versus potentially harmful responses demonstrates that expert evaluators can reliably assess risk-related dimensions, which is critical for deployment in real-world mental health contexts.
    
    \item The variation in agreement across dimensions provides insights into the relative difficulty of different quality dimensions. Guidance showed somewhat lower agreement (91.57\%) compared to other dimensions, suggesting that determining the appropriateness of therapeutic direction may involve more subjective judgment than assessing safety or relevance. Nevertheless, even the lowest agreement rate exceeded 90\%, indicating strong reliability across all evaluated dimensions.
\end{enumerate}

\begin{table*}[!tb]
\centering
\footnotesize
\renewcommand{\arraystretch}{1.3} 
\setlength{\tabcolsep}{3pt} 
\begin{tabularx}{\linewidth}{@{}>{\raggedright\arraybackslash}p{1.8cm}*{7}{>{\centering\arraybackslash\hspace{0pt}}X}@{}}
\toprule
\rowcolor{HeaderBlue}
\textbf{\color{white}Method} &
\textbf{\color{white}Learned judge} &
\textbf{\color{white}Inference-time control} &
\textbf{\color{white}Role-separated refinement} &
\textbf{\color{white}Actionable multi-dim. feedback} &
\textbf{\color{white}Collaborative system} &
\textbf{\color{white}Agentic Roles} &
\textbf{\color{white}Decision refinement} \\
\midrule

\rowcolor{GrayBenchmark}
TherapyGym & Yes & No & No & No & No & No & No \\

\rowcolor{GoldBaseline}
Kardia-R1 & No & No & No & No & No & No & No \\

\rowcolor{BlueHero}
\textbf{Ours} & \textbf{Yes} & \textbf{Yes} & \textbf{Yes} & \textbf{Yes} & \textbf{Yes} & \textbf{Yes} & \textbf{Yes} \\
\bottomrule
\end{tabularx}
\caption{
Precise comparison with closely related therapeutic or supportive-dialogue frameworks. We mark a feature as \textbf{Yes} only when it is an explicit and central component of the method in the same sense used by our formulation. Unlike TherapyGym and Kardia-R1, which primarily use evaluation signals for conventional training-time alignment, our framework separates evaluation from generation by first learning a clinician-aligned evaluator and then operationalizing its structured multi-dimensional judgments as inference-time control signals in an agentic refinement pipeline. This design enables targeted repair of weak or unsafe responses rather than relying solely on policy-level training improvements.
}
\label{tab:positioning_strict_binary}
\end{table*}

\begin{table*}[t]
\centering
\footnotesize
\renewcommand{\arraystretch}{1.15}
\setlength{\tabcolsep}{4pt}
\begin{tabularx}{\textwidth}{p{2.2cm} p{3.4cm} X X X X X}
\toprule
\rowcolor{HeaderBlue}
\textbf{\color{white}Dimension} &
\textbf{\color{white}Definition} &
\textbf{\color{white}5 (Excellent)} &
\textbf{\color{white}4} &
\textbf{\color{white}3} &
\textbf{\color{white}2} &
\textbf{\color{white}1 (Very Poor)} \\
\midrule

\rowcolor{GrayBenchmark}
Guidance &
Clear, structured direction and next steps for the user. &
Provides specific, actionable steps or clear advice. &
Offers mostly clear guidance with minor vagueness. &
Provides general or partially helpful direction. &
Minimal actionable advice; next steps unclear. &
No meaningful guidance or direction. \\

Informativeness &
Usefulness and depth of suggestions or information. &
Very informative with relevant, rich detail. &
Mostly informative with useful insights. &
Moderately informative but somewhat limited. &
Low usefulness; vague or generic suggestions. &
No informative value. \\

\rowcolor{GrayBenchmark}
Relevance &
How well the response addresses the user's specific concern. &
Highly focused and directly addresses the issue. &
Generally on-topic with minor broad content. &
Somewhat related but lacks specificity. &
Mostly off-topic with little relevance. &
Entirely irrelevant or unrelated. \\

Safety &
Compliance with ethical norms and avoidance of harmful suggestions. &
Fully safe and aligns with professional and ethical standards. &
Safe overall but could be more cautious. &
Mostly safe with minor issues or vague cautions. &
Questionable content or missing safeguards. &
Unsafe or ethically concerning. \\

\rowcolor{GrayBenchmark}
Empathy &
Expression of warmth, care, and acknowledgment of emotions. &
Deeply empathic and fully validates the user's feelings. &
Good empathy but could be deeper. &
Moderate emotional understanding. &
Minimal empathy with brief acknowledgment. &
No emotional acknowledgment. \\

Helpfulness &
Contribution to reducing distress and supporting the user emotionally. &
Very reassuring and clearly supportive. &
Helpful and kind but not strongly reassuring. &
Mild encouragement but limited support. &
Indirect help; little emotional support. &
Not helpful or potentially dismissive. \\

\rowcolor{GrayBenchmark}
Understanding &
Accuracy in reflecting the user's emotional or experiential context. &
Clearly reflects and understands the user's situation. &
Good understanding with minor gaps. &
Partial understanding but somewhat misaligned. &
Minimal reflection of user context. &
No evidence of understanding. \\

\bottomrule
\end{tabularx}
\caption{Full seven-dimension therapeutic evaluation rubric used for human and model-based evaluation \cite{badawi2025can}. Each dimension is scored from 1 (Very Poor) to 5 (Excellent) based on the anchor descriptions provided.} 
\label{tab:therapeutic_rubric}

\end{table*}

\begin{table*}[t]
\centering
\footnotesize
\renewcommand{\arraystretch}{1.15}
\setlength{\tabcolsep}{5pt}
\begin{tabularx}{\textwidth}{p{2.6cm} p{3.2cm} X}
\toprule
\rowcolor{HeaderBlue}
\textbf{\color{white}Stage} &
\textbf{\color{white}Clinician Group} &
\textbf{\color{white}Details} \\
\midrule

\rowcolor{GrayBenchmark}
Framework development &
Two licensed mental health clinicians &
Contributed to defining the evaluation dimensions, aligning them with established clinical assessment principles, and refining the scoring criteria. Both had more than 10 years of clinical and academic experience, including a Professor of Psychology and a practicing Psychiatrist affiliated with a mental health institute and hospital. \\

Response evaluation &
Three independent clinical evaluators &
Conducted the blinded human evaluation of responses. All three had graduate-level training in psychology or related mental-health care and direct patient-facing or mental-health service experience. \\

\rowcolor{GrayBenchmark}
Evaluation procedure &
All evaluators &
Completed structured training on the rubric and evaluation protocol before annotation. Responses were anonymized and evaluators were blinded to system identity during rating. \\

\bottomrule
\end{tabularx}
\caption{Clinician involvement in rubric development and blinded response evaluation.}
\label{tab:clinician_details}
\end{table*}

\definecolor{HeaderDark}{RGB}{0, 51, 102}
\definecolor{PurpleGen4}{RGB}{250, 240, 255} 
\definecolor{TealGen8}{RGB}{235, 252, 250}   
\definecolor{BlueGen16}{RGB}{232, 242, 255}  

\begin{table*}[!tb]
\centering
\small 
\renewcommand{\arraystretch}{1.2}
\begin{tabularx}{\textwidth}{l X S[table-format=1.3] l S[table-format=1.3] S[table-format=1.3]}
\toprule
\rowcolor{HeaderDark}
\textbf{\color{white}Judge Group} & \textbf{\color{white}Dimension} & {\textbf{\color{white}ICC(C,1)}} & \textbf{\color{white}95\% CI} & {\textbf{\color{white}ICC(A,1)}} & {\textbf{\color{white}CI width}} \\
\midrule

\rowcolor{PurpleGen4} & Guidance         & 0.980 & [0.883, 0.996] & 0.977 & 0.113 \\
\rowcolor{PurpleGen4} & Informativeness  & 0.977 & [0.962, 0.998] & 0.966 & 0.036 \\
\rowcolor{PurpleGen4} & Relevance        & 0.953 & [0.825, 0.993] & 0.920 & 0.168 \\
\rowcolor{PurpleGen4} & Safety           & \textbf{0.896} & [0.290, 0.982] & 0.842 & 0.952 \\
\rowcolor{PurpleGen4} & Empathy          & 0.884 & [0.415, 0.937] & 0.827 & 0.522 \\
\rowcolor{PurpleGen4} & Helpfulness      & 0.857 & [0.376, 0.928] & 0.833 & 0.552 \\
\rowcolor{PurpleGen4} \multirow{-7}{*}{TheraJudge (gen 4)} & Understanding & 0.920 & [0.425, 0.996] & 0.787 & 0.570 \\
\midrule

\rowcolor{TealGen8} & Guidance         & 0.948 & [0.667, 0.984] & 0.938 & 0.317 \\
\rowcolor{TealGen8} & Informativeness  & \textbf{0.988} & [0.966, 0.995] & 0.989 & 0.030 \\
\rowcolor{TealGen8} & Relevance        & 0.927 & [0.686, 0.986] & 0.854 & 0.300 \\
\rowcolor{TealGen8} & Safety           & 0.835 & [0.144, 0.932] & 0.625 & 0.918 \\
\rowcolor{TealGen8} & Empathy          & 0.808 & [0.068, 0.900] & 0.745 & 0.832 \\
\rowcolor{TealGen8} & Helpfulness      & 0.866 & [0.250, 0.951] & 0.817 & 0.701 \\
\rowcolor{TealGen8} \multirow{-7}{*}{TheraJudge (gen 8)} & Understanding & 0.930 & [0.653, 0.990] & 0.880 & 0.337 \\
\midrule

\rowcolor{BlueGen16} & Guidance         & \bfseries 0.989 & [0.948, 0.998] & 0.981 & 0.050 \\
\rowcolor{BlueGen16} & Informativeness  & 0.983 & [0.918, 0.995] & 0.977 & 0.077 \\
\rowcolor{BlueGen16} & Relevance        & \bfseries 0.960 & [0.900, 0.985] & 0.929 & 0.085 \\
\rowcolor{BlueGen16} & Safety           & 0.879 & [0.561, 0.958] & 0.848 & 0.901 \\
\rowcolor{BlueGen16} & Empathy          & \bfseries 0.932 & [0.709, 0.979] & 0.919 & 0.270 \\
\rowcolor{BlueGen16} & Helpfulness      & \bfseries 0.945 & [0.694, 0.986] & 0.934 & 0.291 \\
\rowcolor{BlueGen16} \multirow{-7}{*}{\textbf{TheraJudge (gen 16)}} & Understanding & \bfseries 0.960 & [0.882, 0.987] & 0.946 & 0.106 \\
\bottomrule
\end{tabularx}
\caption{Performance variation with increasing number of generations per prompt in GRPO. Gen 4, Gen 8, and Gen 16 denote the number of response candidates sampled per prompt during GRPO preference optimization.}
\label{tab:grpo_iterations_comparison}
\end{table*}

\begin{table*}[t]
\centering
\tiny
\renewcommand{\arraystretch}{1.1}
\setlength{\tabcolsep}{4pt}
\begin{tabular}{@{}llcccccc@{}}
\toprule
\textbf{Judge} & \textbf{Dimension} & \textbf{ICC(C,1)} & \textbf{95\% CI} & \textbf{ICC(A,1)} & \textbf{CI Width} & \textbf{Status} & \textbf{Rank} \\
\midrule
\multirow{7}{*}{\textbf{TheraJudge (gen 16)}}
& Guidance        & 0.989 & [0.948, 0.998] & 0.981 & 0.050 & Excellent  & 1/7 \\
& Informativeness & 0.983 & [0.918, 0.995] & 0.977 & 0.077 & Excellent  & 1/7 \\
& Relevance       & 0.960 & [0.900, 0.985] & 0.929 & 0.085 & Excellent  & 1/7 \\
& Safety          & 0.879 & [0.561, 0.958] & 0.848 & 0.901 & Poor$^*$   & 1/7 \\
& Empathy         & 0.932 & [0.709, 0.979] & 0.919 & 0.270 & Excellent  & 1/7 \\
& Helpfulness     & 0.945 & [0.694, 0.986] & 0.934 & 0.291 & Excellent  & 1/7 \\
& Understanding   & 0.960 & [0.882, 0.987] & 0.946 & 0.106 & Excellent  & 1/7 \\
\cmidrule{2-8}
& \textbf{Average} & \textbf{0.949} & -- & \textbf{0.933} & \textbf{0.254} & \textbf{7/7 Excellent} & \textbf{1st} \\
\midrule
\multirow{7}{*}{\textbf{Qwen-2.5-7B-SFT}}
& Guidance        & 0.929 & [0.852, 0.966] & 0.911 & 0.114 & Excellent  & 3/7 \\
& Informativeness & 0.895 & [0.805, 0.983] & 0.886 & 0.178 & Excellent  & 4/7 \\
& Relevance       & 0.760 & [0.636, 0.811] & 0.695 & 0.175 & Excellent  & 2/7 \\
& Safety          & 0.605 & [0.429, 0.672] & 0.451 & 0.242 & Excellent  & 3/7 \\
& Empathy         & 0.874 & [0.459, 0.978] & 0.866 & 0.518 & Moderate   & 4/7 \\
& Helpfulness     & 0.921 & [0.820, 0.980] & 0.926 & 0.160 & Excellent  & 2/7 \\
& Understanding   & 0.803 & [0.646, 0.862] & 0.743 & 0.216 & Excellent  & 5/7 \\
\cmidrule{2-8}
& \textbf{Average} & \textbf{0.827} & -- & \textbf{0.783} & \textbf{0.229} & \textbf{6/7 Excellent} & \textbf{2nd} \\
\midrule
\multirow{7}{*}{\textbf{Claude-3.7-Sonnet}}
& Guidance        & 0.881 & [0.764, 0.980] & 0.837 & 0.216 & Excellent  & 4/7 \\
& Informativeness & 0.915 & [0.830, 0.972] & 0.915 & 0.142 & Excellent  & 3/7 \\
& Relevance       & 0.730 & [0.394, 0.987] & 0.743 & 0.594 & Poor       & 3/7 \\
& Safety          & 0.685 & [0.333, 0.961] & 0.597 & 0.628 & Poor       & 2/7 \\
& Empathy         & 0.906 & [0.429, 0.958] & 0.474 & 0.528 & Moderate   & 2/7 \\
& Helpfulness     & 0.900 & [0.734, 0.992] & 0.742 & 0.258 & Excellent  & 3/7 \\
& Understanding   & 0.791 & [0.563, 0.956] & 0.806 & 0.394 & Moderate   & 6/7 \\
\cmidrule{2-8}
& \textbf{Average} & \textbf{0.830} & -- & \textbf{0.731} & \textbf{0.394} & \textbf{5/7 Excellent} & \textbf{3rd} \\
\midrule
\multirow{7}{*}{\textbf{o4-mini}}
& Guidance        & 0.948 & [0.744, 0.976] & 0.786 & 0.233 & Excellent  & 2/7 \\
& Informativeness & 0.918 & [0.638, 0.978] & 0.908 & 0.340 & Excellent  & 2/7 \\
& Relevance       & 0.342 & [0.069, 0.673] & 0.140 & 0.605 & Poor       & 6/7 \\
& Safety          & 0.259 & [0.081, 0.703] & 0.117 & 0.621 & Poor       & 6/7 \\
& Empathy         & 0.883 & [0.476, 0.945] & 0.499 & 0.469 & Moderate   & 3/7 \\
& Helpfulness     & 0.871 & [0.578, 0.934] & 0.660 & 0.356 & Moderate   & 4/7 \\
& Understanding   & 0.871 & [0.636, 0.938] & 0.592 & 0.302 & Excellent  & 2/7 \\
\cmidrule{2-8}
& \textbf{Average} & \textbf{0.727} & -- & \textbf{0.529} & \textbf{0.418} & \textbf{5/7 Excellent} & \textbf{4th} \\
\midrule
\multirow{7}{*}{\textbf{GPT-4o}}
& Guidance        & 0.849 & [0.650, 0.975] & 0.475 & 0.324 & Excellent  & 6/7 \\
& Informativeness & 0.856 & [0.655, 0.964] & 0.681 & 0.310 & Excellent  & 6/7 \\
& Relevance       & 0.532 & [0.267, 0.826] & 0.243 & 0.559 & Moderate   & 4/7 \\
& Safety          & 0.480 & [0.116, 0.858] & 0.279 & 0.741 & Poor       & 4/7 \\
& Empathy         & 0.835 & [0.331, 0.891] & 0.288 & 0.560 & Moderate   & 6/7 \\
& Helpfulness     & 0.800 & [0.407, 0.924] & 0.457 & 0.517 & Moderate   & 5/7 \\
& Understanding   & 0.823 & [0.549, 0.884] & 0.485 & 0.334 & Excellent  & 4/7 \\
\cmidrule{2-8}
& \textbf{Average} & \textbf{0.739} & -- & \textbf{0.415} & \textbf{0.478} & \textbf{5/7 Excellent} & \textbf{5th} \\
\midrule
\multirow{7}{*}{\textbf{Qwen-2.5-7B-ZS}}
& Guidance        & 0.650 & [0.450, 0.824] & 0.499 & 0.374 & Moderate   & 7/7 \\
& Informativeness & 0.802 & [0.573, 0.949] & 0.669 & 0.376 & Moderate   & 7/7 \\
& Relevance       & 0.519 & [0.310, 0.602] & 0.263 & 0.292 & Excellent  & 5/7 \\
& Safety          & 0.145 & [0.155, 0.441] & 0.079 & 0.427 & Moderate   & 7/7 \\
& Empathy         & 0.616 & [0.164, 0.708] & 0.602 & 0.544 & Moderate   & 7/7 \\
& Helpfulness     & 0.761 & [0.394, 0.887] & 0.740 & 0.493 & Moderate   & 6/7 \\
& Understanding   & 0.870 & [0.471, 0.924] & 0.712 & 0.453 & Moderate   & 3/7 \\
\cmidrule{2-8}
& \textbf{Average} & \textbf{0.623} & -- & \textbf{0.509} & \textbf{0.423} & \textbf{3/7 Excellent} & \textbf{6th} \\
\midrule
\multirow{7}{*}{\textbf{Gemini-2.5-Flash}}
& Guidance        & 0.855 & [0.557, 0.956] & 0.682 & 0.398 & Moderate   & 5/7 \\
& Informativeness & 0.878 & [0.522, 0.962] & 0.877 & 0.439 & Moderate   & 5/7 \\
& Relevance       & 0.306 & [0.011, 0.767] & 0.137 & 0.755 & Poor       & 7/7 \\
& Safety          & 0.377 & [0.047, 0.868] & 0.222 & 0.790 & Poor       & 5/7 \\
& Empathy         & 0.838 & [0.401, 0.918] & 0.380 & 0.517 & Moderate   & 5/7 \\
& Helpfulness     & 0.734 & [0.271, 0.832] & 0.385 & 0.561 & Poor       & 7/7 \\
& Understanding   & 0.362 & [0.137, 0.781] & 0.180 & 0.644 & Poor       & 7/7 \\
\cmidrule{2-8}
& \textbf{Average} & \textbf{0.621} & -- & \textbf{0.409} & \textbf{0.586} & \textbf{3/7 Excellent} & \textbf{7th} \\
\bottomrule
\end{tabular}
\caption{\small{Comprehensive judge comparison with complete rankings. All ranks computed by comparing ICC(C,1) values across all 7 judges for each dimension. $^*$Poor status due to wide CI despite high ICC. TheraJudge (gen 16) achieves best overall performance with universal excellent reliability and wins 5/7 dimension rankings.}}
\label{tab:comprehensive_judge_comparison}
\end{table*}

\end{document}